\documentclass[letterpaper]{article} % DO NOT CHANGE THIS
\usepackage{aaai23}  % DO NOT CHANGE THIS
\usepackage{times}  % DO NOT CHANGE THIS
\usepackage{helvet}  % DO NOT CHANGE THIS
\usepackage{courier}  % DO NOT CHANGE THIS
\usepackage[hyphens]{url}  % DO NOT CHANGE THIS
\usepackage{graphicx} % DO NOT CHANGE THIS
\usepackage{natbib}  % DO NOT CHANGE THIS AND DO NOT ADD ANY OPTIONS TO IT
\usepackage{caption} % DO NOT CHANGE THIS AND DO NOT ADD ANY OPTIONS TO IT
\usepackage{algorithm}
\usepackage{algorithmic} 

\usepackage{newfloat}
\usepackage{listings}
\DeclareCaptionStyle{ruled}{labelfont=normalfont,labelsep=colon,strut=off} % DO NOT CHANGE THIS
\floatstyle{ruled}
\newfloat{listing}{tb}{lst}{}
\floatname{listing}{Listing}
\title{GRPE: Relative Positional Encoding for Graph Transformer}
\author{
\textsuperscript{\rm 1}Wonpyo Park\equalcontrib, \textsuperscript{\rm 2}Woonggi Chang\equalcontrib, \textsuperscript{\rm 2}Donggeon Lee, \textsuperscript{\rm 2}Juntae Kim, \textsuperscript{\rm 1}Seung-won Hwang
}
\affiliations{
    \textsuperscript{\rm 1}Seoul National University, \textsuperscript{\rm 2}Standigm
}

\newcommand{\eg}{\textit{e.g.}}
\newcommand{\ie}{\textit{i.e.}}

\usepackage{bibentry}
\usepackage{amsmath} % for the equation* environment
\usepackage{amsfonts} 
\usepackage{booktabs}
\usepackage{subcaption}

\begin{document}

\maketitle
%%%%%%%%% ABSTRACT
\begin{abstract}

% Designing an efficient model to encode graphs is a key challenge of molecular representation learning.
We propose a novel positional encoding for learning graph on Transformer architecture.
% Our method considers relative relational information between nodes to incorporate grap
% Transformer built upon self-attention is a natural choice for graph processing, but it requires explicit incorporation of positional information.
% Self-attention is invariant to the order of the inputs, thus it requires explicit incorporation of positional information.
% Designing an efficiently representation of position of each node in a graph is key challenge for learning graph Transformer.
% Transformer built upon self-attention requires explicit incorporation of positional information of inputs.
Existing approaches either linearize a graph to encode absolute position in the sequence of nodes, or encode relative position with another node using bias terms.
The former loses preciseness of relative position from linearization, while the latter loses a tight integration of node-edge and node-topology interaction. 
% In this work, we propose relative positional encoding for a graph to overcome the weakness of the previous approaches.
To overcome the weakness of the previous approaches, our method encodes a graph without linearization and considers both node-topology and node-edge interaction.
We name our method Graph Relative Positional Encoding dedicated to graph representation learning.
Experiments conducted on various graph datasets show that the proposed method outperforms previous approaches significantly.
% Our code is publicly available at \texttt{https://github.com/lenscloth/GRPE}.
\end{abstract}
\section{Introduction}
\label{sec:intro}

Transformer \cite{vaswani2017attention} built upon a self-attention module is permutation equivariant where an order of inputs does not affect corresponding outputs.
Therefore, Transformer requires explicit representations of the position of inputs to effectively learn any structured data. 
In the case of natural language processing or computer vision, absolute positional encoding is widely adopted as each input has its own absolute position, \eg, $n$-th order of word in a sentence or coordinates of a patch within a grid.
The positional embedding vector of the absolute position is added to the input before feeding to Transformer.
% In the case of natural language processing or computer vision, each input has its absolute position, \eg, $n$-th order of word in a sentence or coordinates of a patch within a grid.
% In general, an embedding vector corresponding to the input's absolute position is added to each input embedding before feeding to Transformer.
% For example, the original work \cite{vaswani2017attention} leverages sinusoidal positional encoding.
However, adopting this absolute positional encoding on a graph is not trivial since nodes do not have absolute positions such as order.
Therefore, representing the position of each node in a graph is a key challenge in designing a Transformer for a graph.

Several works have been proposed to incorporate positional information of graph on Transformer, and we
categorize existing works into two: (a) linearizing graph to encode the absolute position of each node~\citep{gt_dwivedi2020generalization,san_kreuzer2021rethinking} using techniques like graph Laplacian or singular value decomposition (b) encoding position relative to another node with bias terms~\citep{ying2021transformers}. 
The former loses precision of position due to linearization, while the latter loses a tight integration of node-edge and node-topology information.

% Our method captures relative relations between two nodes such as the shortest path distance. 
% % Figure~\ref{fig:teaser} shows how our method reflects topological information on attention map, incorporating node-topology information.

%%%%%%%%%%%%%%%% front page figure %%%%%%%%%%%%%%%%
\begin{figure}[t]
\centering\includegraphics[width=2.2\linewidth]{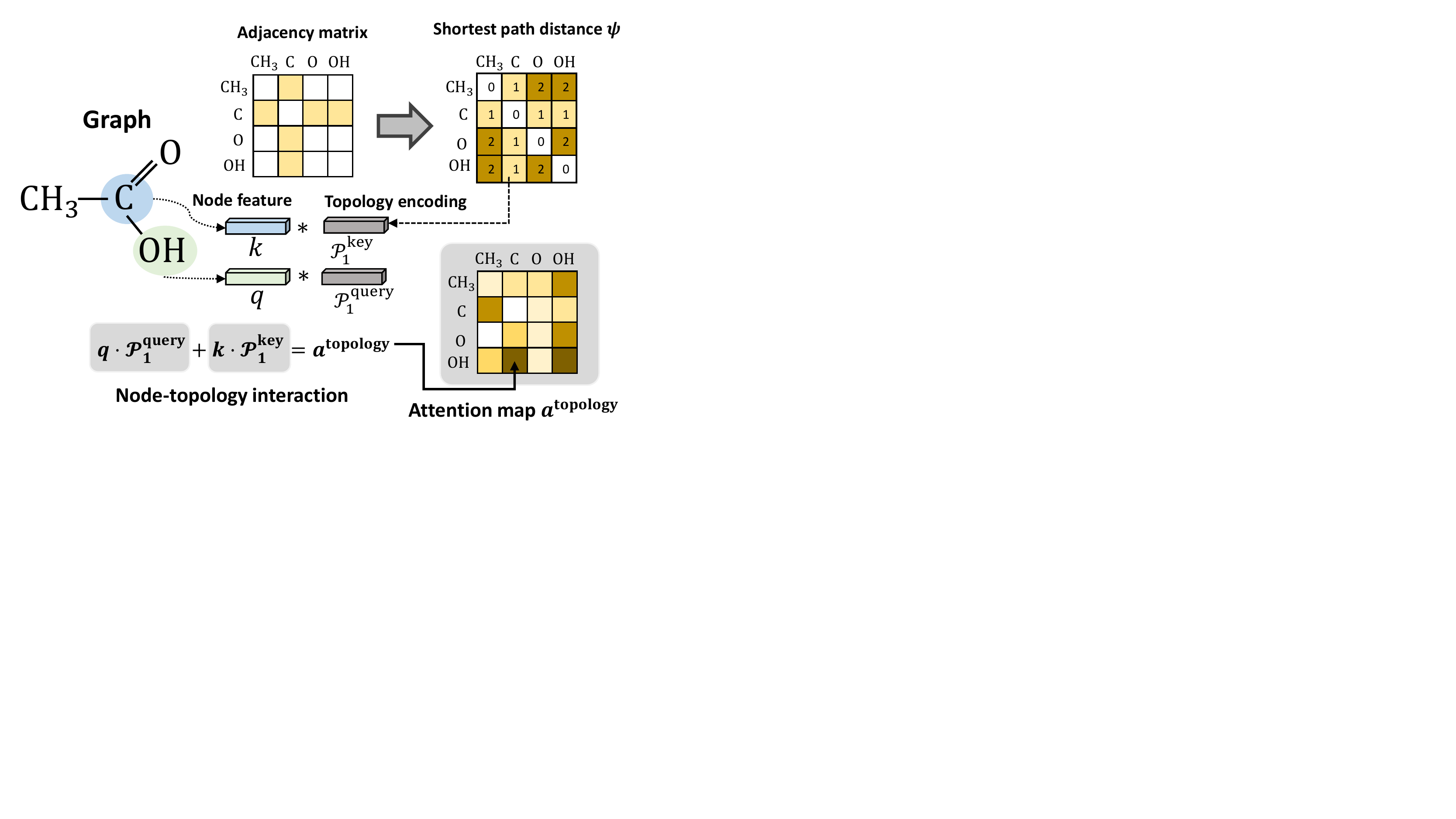}
\vspace{-5.2cm}
\caption{
% Our proposed Graph Relative Positional Encoding (GRPE). 
% % GSA is a variant of self-attention module of Transformer~\citep{vaswani2017attention} to learn graph representation.
% % GSA learns graph representation by encoding graph structure on both scaled dot product attention map. and the hidden representation.
% GRPE is a relative positional encoding method dedicated to learn graph representation.
% How our method builds an attention map. 
% The shortest path distance between two nodes is used to represent the topology of a graph.
% % Node feature and topology encoding vector interact to incorporate node-topology information on the attention map.
% Node-topology information is the interaction between node feature and topology encoding vector. 
% It is encoded on the attention map along with node-edge information which is the interaction between node feature and edge encoding vector. 
% Therefore, any pairs of nodes with the same topological relation can have diverse attention maps according to node type.
% We skip the illustration of node-edge information for simplification.
How our method builds attention map $a^\text{topology}$ reflcting node-topology information, by interaction between node feature $q,k$ and topology encoding $\mathcal{P}$ representing adjacency and shortest path.
We illustrate with $\mathcal{P}_{1}$ capturing
relations within 1-hop distance, though we generalize for $\mathcal{P}_{l}$.
Similarly, attention map $a^\text{edge}$ reflects diverse edge types, and interaction between node feature and edge encoding.
 We skip the illustration of node-edge information for simplification.
}
\label{fig:teaser}
\end{figure}
%%%%%%%%%%%%%%%%%%%%%%%%%%%%%%%%%%%%%%%%%%%%%%%%%%%

%%%%%%%%%%%%%%%% front page figure %%%%%%%%%%%%%%%%
\begin{figure*}[t]
\centering\includegraphics[width=1\linewidth]{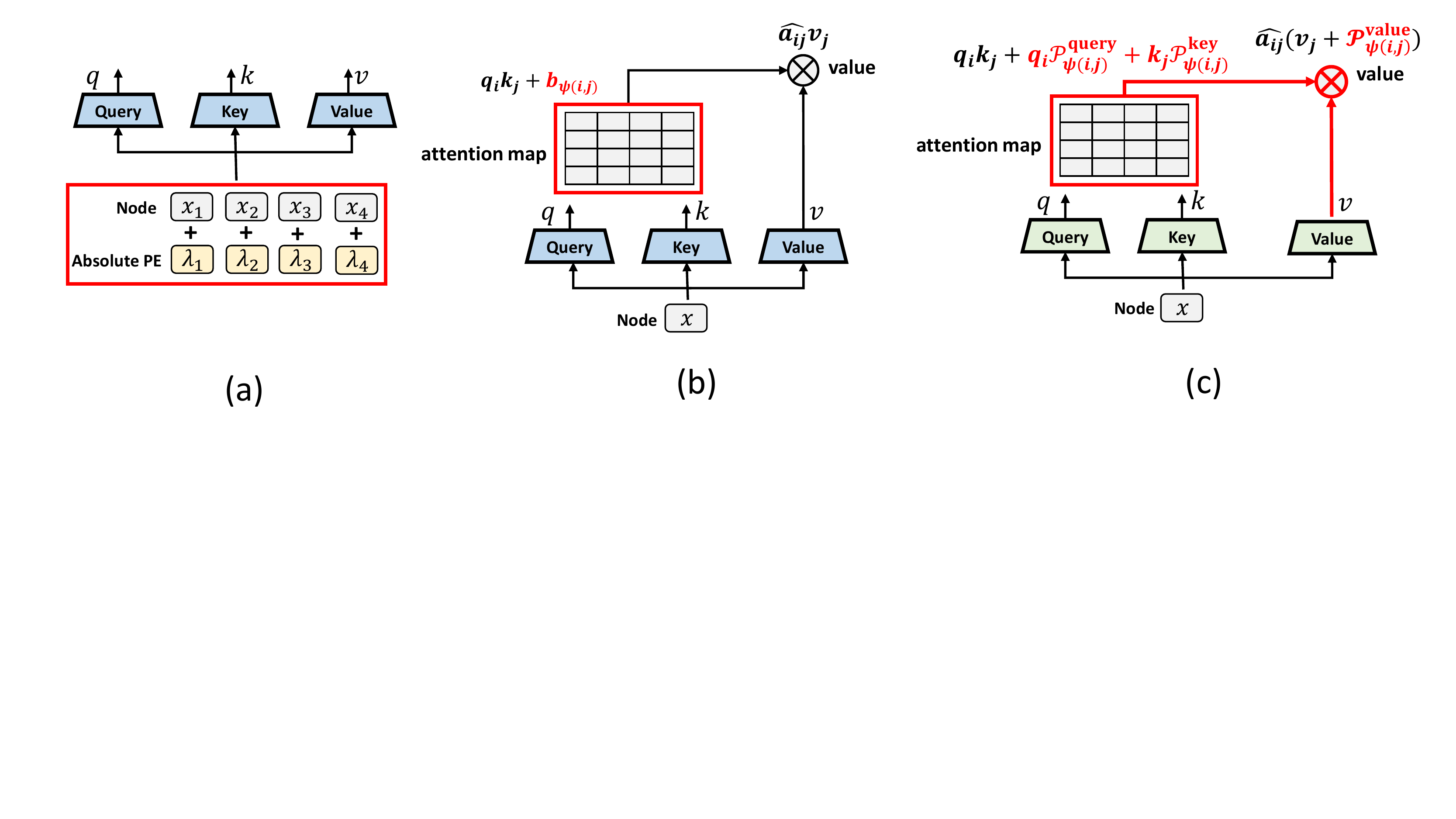}
\vspace{-5.2cm}
\caption{
% % Our proposed Graph Relative Positional Encoding (GRPE). 
% % % GSA is a variant of self-attention module of Transformer~\citep{vaswani2017attention} to learn graph representation.
% % % GSA learns graph representation by encoding graph structure on both scaled dot product attention map. and the hidden representation.
% % GRPE is a relative positional encoding method dedicated to learn graph representation.
% Comparison between previous approaches and our GRPE. Red colored components indicate where graph positional information is encoded in self-attention. 
% Existing work linearize a graph to obtain absolute positional encoding for each node or add bias terms on the attention map. 
% On the other hand, our method considers node-topology and node-edge relations when building an attention map, and incorporates graph on the hidden representations of value. 
% Likewise, our method encodes a graph on both an attention map and value.
% Note that, we skip the illustration of edge encoding for simplification.
Contrasting ours from previous approaches by where graph positional information is encoded, \ie, components marked in red.
% Existing works either require (a) linearization or 
Existing works either (a) linearize a graph to obtain absolute positional encoding for initial inputs or (b) add bias terms for attention map. 
On the other hand, \textbf{(c) our method} encodes on both attention map and value, capturing node-topology and node-edge relations.
We skip the illustration of edge encoding for simplification.
}
\label{fig:figure2}
\end{figure*}
%%%%%%%%%%%%%%%%%%%%%%%%%%%%%%%%%%%%%%%%%%%%%%%%%%%

On the other hand, ours can be interpreted as overcoming the weakness of (a) and (b):
 Figure~\ref{fig:teaser} illustrates that our method reflects global topology as attention map, while incorporating node-topology interaction.
More specifically, unlike conventional approaches (a) and (b),  limited to encode graph only on either the initial input (Figure~\ref{fig:figure2}a) or attention map (Figure~\ref{fig:figure2}b),
our method  encodes positional information when node features interact with each other on self-attention.
%instead of adding positional embedding on initial inputs.
Figure~\ref{fig:figure2}c shows our architecture where the topology and edge of a graph are encoded on both attention map and value.
% Figure~\ref{fig:figure2} shows our architecture where the topology and edge of a graph are encoded on both attention map and value.

To this end, we introduce two set of learnable positional encoding vectors which represent relative positional relation.
% Thus, our method does not require linearization of a graph.
% To this end, we introduce two sets of learnable positional encoding vectors.
The first is topology encoding to represent topological relation between nodes.
The second is edge encoding to represent connection between nodes.
% Instead of adding positional embedding on initial inputs, our method encode positional information when node features interact with each other on self-attention.
Node features and the two encoding vectors interact to integrate both node-topology and node-edge interaction when building attention map.
% Our method considers the interaction between node features and the two encoding vectors to integrate both node-spatial relation and node-edge information when building attention map.
Furthermore, we leverage the two positional encodings to incorporate graph on the hidden representations of self-attention.
With these two positional encodings, topology of a graph can be represented in both attention map and value.
% To avoid losing relative position due to the linearization, we propose to adopt the relative positional encoding by \cite{shaw2018self}.
% Its efficacy has been verified in natural language processing, however, the efficacy is not yet studied in a graph.
% Since the original work is designed for 1D word sequences, we reformulate to incorporate graph-specific properties.
% To this end, 
% Our method considers the interaction between node features and the two encoding vectors to integrate both node-spatial relation and node-edge information.

% To avoid losing relative position due to the linearization, we propose a novel relative positional encoding method for graph. 
% To overcome the weakness of the two, we propose a novel positional encoding for graph which considers both relative position and the interaction with nodes.
% To avoid losing positional information on a graph due to the linearization, we propose a relative positional encoding for a graph.
In terms of utilizing relative position between inputs and incorporating interaction between input feature and positional encoding, our work is related to the work of \citeauthor{shaw2018self}.
Its efficacy has been verified in sequences, e.g., natural language processing, but does not generalize to  a graph. 
% %Since the original work is designed for 1D word sequences, we reformulate to incorporate graph-specific properties. 
% The original work is designed 
% to incorporate graph-specific properties. 
Our proposed relative positional encoding generalizes from 1D sequence structure to relative positions in the graph,
 dedicated to graph representation learning covering graph-specific properties.

We name our method for graph representation learning as \textbf{G}raph \textbf{R}elative \textbf{P}ositional \textbf{E}ncoding (GRPE).
We extensively conducted experiments to validate the efficacy of our proposed method on various tasks, \eg, graph classification, graph regression and node classification.
Models built with our relative positional encoding achieve state-of-the-art performance on various graph datasets, showing the efficacy of our proposed method.

\section{Related Work}
\label{sec:related_work}

% \subsection{Graph Convolutional Network} 

% Graph Convolutional Network (GCN)~\cite{gcn_kipf2016semi,velivckovic2017graph,mpnn_gilmer2017neural,gin_xu2018powerful,pna_corso2020principal} approaches are based on the message passing scheme, which recursively aggregates information from neighboring nodes. GCN approaches are known to suffer from suspended animation problem~\cite{zhang2019gresnet} and over-smoothing problem~\cite{li2018deeper, xu2018representation,zhao2019pairnorm} when stacking multiple layers.
% According to ~\cite{zhang2019gresnet}, as GCN-based model stacks multiple layers more than a certain limit, the model will not respond to training data causing the suspended animation problem.
% Meanwhile, over-smoothing problem which the performance degrades as representations of all the nodes become indistinguishable. These two issues greatly hinder the application of GCNs for deep graph representation learning tasks although several approaches~\cite{li2018deeper,deepergn_li2020deepergcn} tackle these problems. Transformer architecture is relatively free from those issues with stacking multiple layers.

% \subsection{Transformer for Graph}

% Many attempts leverage Transformer architecture to learn graph representation.
% \cite{velivckovic2017graph} apply self-attention mechanism on graph convolutional network where self-attention is only applied within neighboring nodes.

Existing works leverage Transformer architecture to learn graph representation. We categorize those methods as follows.

% (1) adopting Transformer without positional information (2) linearizing graph to encode absolute position of each node (3) encoding relative position to another node.

% can be categorized as follow.

% Early model adopts Transformer iwthout position infformation for graph repersentation.

% Absolute positional..
Earlier models adopt Transformer without explicit encoding of positional information on a graph.
% Transformer architecture are employed in graph convolutional networks. 
\citeauthor{velivckovic2017graph}  replace graph convolution operation with self-attention module where attention
is only performed within neighboring nodes.
\citeauthor{rong2020self} stack self-attention module next to the graph convolutional networks iteratively to consider long-range interaction between nodes.
In their method, affinity is considered only on the graph convolutional networks, and positional information is not given on self-attention.

Later works employ absolute positional encoding to explicitly encode positional information of graph on Transformer.
% Absolute positional encoding is adopted to linearize a graph into a sequence.
% Some works linearize a graph to sequence to employ absolute positional encoding.
Their main idea is to linearize a graph into a sequence of nodes, and an absolute positional encoding is added to the input feature.
\citeauthor{gt_dwivedi2020generalization} adopted graph Laplacian as a positional encoding, where each cell of encoding represents partitions after graph min-cut. 
Nodes sharing many partitions after graph min-cut would have similar graph Laplacian vectors.
% Absolute positional encoding for graph is devised to encode graph within the input features.
% \cite{gt_dwivedi2020generalization} adopted graph Laplacian as a positional encoding, where each cell of encoding represents partitions after graph min-cut. 
% Nodes sharing many partitions after graph min-cut would have similar graph Laplacian 
% vectors.
\citeauthor{san_kreuzer2021rethinking} employ a learnable positional encoding with a Transformer where its input is the Laplacian spectrum of a graph. Due to the linearization of a graph, those approaches lose the preciseness of position on the graph.
% On the other hand, our method encode graph on self-attention without linearization  and relative positional information is encoded without loss.
% The models of our method achieve better performance on molecule property prediction benchmarks compared to the methods utilizing absolute positional encoding for graph.

Meanwhile, encoding relative positional information has been studied to avoid losing the preciseness of position.
Graphormer introduced by \citeauthor{ying2021transformers} encodes relative position on scaled dot product attention map by adding bias terms. 
% encodes relative position on scaled dot product attention map by adding bias terms. %. such as the shortest path distance between two nodes.
% They define relative position between nodes with edge and spatial relation, \eg, shortest path distance.
However, the bias terms are parameterized only relative position such as shortest path distance or edge type, and the interaction with node features is lost.
On the other hand, \citeauthor{shaw2018self} introduce relative positional encoding, 
%which considers tight integration of node and relative position, however, their method is designed to process 1D word sequences.
%We extend the systematic design of \citeauthor{shaw2018self} for a graph. Our method considers
for 1D sequence, on which we add relative position encoding, to capture
the interaction between nodes and graph-specific properties such as edge and topology.

\section{Background}
\label{sec:background}

\smallbreak
\subsection{Notation}
We denote a set of nodes on the graph  $\{n_i\}_{i=1:N}$ and a set of edges on the graph $\{e_{ij} |  j\in\mathcal{N}_i\}_{i=1:N}$, where $N$ is the number of nodes and 
$\mathcal{N}_i$ is a set neighbors of a node $n_i$. 
Both $n_i$ and $e_{ij}$ are positive integer numbers to index the type of nodes or edges, \eg, atom numbers or bond types of a molecule.
$\psi(i, j)$ denotes a function encodes topological relationship between the node $n_i$ and $n_j$. 
% For the remaining section, $\psi$ is a structural relationship encoder that outputs the shortest path distance between node $n_i$ and $n_j$. 
% `$\cdot$' denotes the inner product between two vectors.

\subsection{Self-attention} 
Transformer is built by stacking multiple self-attention layers.
Self-attention maps a query and a set of key pairs to compute an attention map. 
Values are weighted summed with the weight on the attention map to output the hidden feature for the following layer.

Specifically, $x_i\in\mathbb{R}^{d_x}$ denotes the input feature of the node $n_i$, and $z_i\in\mathbb{R}^{d_z}$ denotes the output feature of the self-attention module.
The self-attention module computes query $q$, key $k$, and value $v$ with independent linear transformations: $W^{\text{query}}\in\mathbb{R}^{d_x{\times}d_z}$, $W^{\text{key}}\in\mathbb{R}^{d_x{\times}d_z}$ and $W^{\text{value}}\in\mathbb{R}^{d_x{\times}d_z}$.
\begin{equation}
\label{eq:key_query_value}
    q =W^{\text{query}}x\text{,} \quad k = W^{\text{key}}x \quad\text{and}\quad v = W^{\text{value}}x.
\end{equation}
The attention map is computed by applying a scaled dot product between the queries and the keys.
\begin{equation}
\label{eq:tranformer_attention}
a_{ij} = \frac{q_i{\cdot}k_j}{\sqrt{d_z}} \quad \text{and} \quad \hat{a}_{ij} = \frac{\text{exp}(a_{ij})}{\sum_{k=1}^{N}\text{exp}(a_{ik})}.
\end{equation}
The self-attention module outputs the next hidden feature by applying weighted summation on the values.
\begin{equation}
\label{eq:transformer_value}
z_i = \sum_{j=1}^{N}\hat{a}_{ij}v_j.
\end{equation}
$z$ is later fed into a feed forward neural network with a residual connection \citep{he2016deep}. However, we defer detailed explanations since it is out of the scope of our paper.
In practice, self-attention module with multi-head is adopted.

\subsection{Graph with Transformer}

To encode graph topology in Transformer, previous methods focus on encoding graph information into either the attention map or input features fed to Transformer.
% To encode graph structure in the transformer layer, Graphormer 
% One approach is to encode in the self-attention map $a$.
Graphormer \citep{ying2021transformers} adopted two additional terms on the self-attention module to encode graph information on the attention map.
% : (1) spatial encoding with learnable scalar bias (2) edge encoding with learnable scalr bias.
\begin{equation}
\label{eq:graphormer_attention}
a^{\text{Graphormer}}_{ij} = \frac{q_i{\cdot}k_j}{\sqrt{d_z}} + b_{\psi(i,j)} +  \mathcal{E}_{e_{ij}}\cdot{w}.
\end{equation}

$\psi$ represents the topological relation between $n_i$ and $n_j$, which outputs the shortest path distance between the two nodes.
% `$\cdot$' denotes the inner product between two vectors.
Learnable scalar bias $b_{\psi(i,j)}$ encodes topological relation between two nodes, \eg, $b_l$ is a bias representing two nodes that are $l$-hop apart.
% $\mathcal{E}_{e_{ij}}$ and $w$ are learnable embeddings to encode edge information between nodes.
% When two matrcies are givne $\mathcal{E}\in\mathbb{R}^{E\times{d_\mathcal{E}}}$ and $w\in\mathbb{R}^{E\times{d_\mathcal{E}}}$.
% $\mathcal{E}_K{\cdot}{w}_K$ indicates the type of the edge, and it indexes the two matrices $\mathcal{E}\in\mathbb{R}^{E\times{d_\mathcal{E}}}$ and $w\in\mathbb{R}^{E\times{d_\mathcal{E}}}$. 
An embedding vector $\mathcal{E}_{e_{ij}}$ is a feature representing edge between the node $n_i$ and the node $n_j$, and $w$ is a learnable vector.
$\mathcal{E}_{e_{ij}}\cdot{w}$ encodes edge between the two nodes.
Moreover, Graphormer adds centrality encoding into the input $x$ which represents the number of edges of a node.
However, Graphormer encodes graphs on the attention map without considering node-topology and node-edge interaction, on the other hand GRPE considers the two.
We will explain the details in later.

\citeauthor{gt_dwivedi2020generalization} and \citeauthor{san_kreuzer2021rethinking}  utilize graph Laplacian \citep{belkin2003laplacian} $\lambda\in\mathbb{R}^{|N|{\times}d_x}$ as positional encodings on the input feature $x$; $\lambda$ are the top-$d_{x}$ smallest eigenvectors of $I - D^{-\frac{1}{2}}AD^{-\frac{1}{2}}$ where $I$ is an identity matrix, $A$ is an adjacency matrix and $D$ is a degree matrix.
% The graph Laplacian $\lambda_i$ represents the structure of a graph with respect to node $n_i$.
Each cell of a graph Laplacian vector represents partitions after graph min-cut, and neighbouring nodes sharing the many partitions would have similar graph Laplacian.
The graph Laplacian $\lambda_i$ represents the topology of a graph with respect to node $n_i$.

\begin{equation}
\label{eq:laplacian_pe}
\hat{x_i} = x_i + \lambda_i.
\end{equation}

\citeauthor{san_kreuzer2021rethinking} adopt an additional Transformer model $f$ to produce learnable positional encoding: $\hat{x}_i =x_i + \hat{\lambda}_i$ where $\hat{\lambda}=f(\lambda)$.
By adding the graph Laplacian into input $x$, graph information can be encoded in both the attention map and the hidden representations.
Their methods lose relative positional information during the linearization of a graph to obtain absolute positional encoding.
However, our method encodes a graph directly on the attention map without linearization, thus relative positional information is encoded without loss.
\section{Our Approach}
\label{sec:method}

\begin{figure}[t]
\hspace{4.7cm}\centerline{\includegraphics[width=2.2\linewidth]{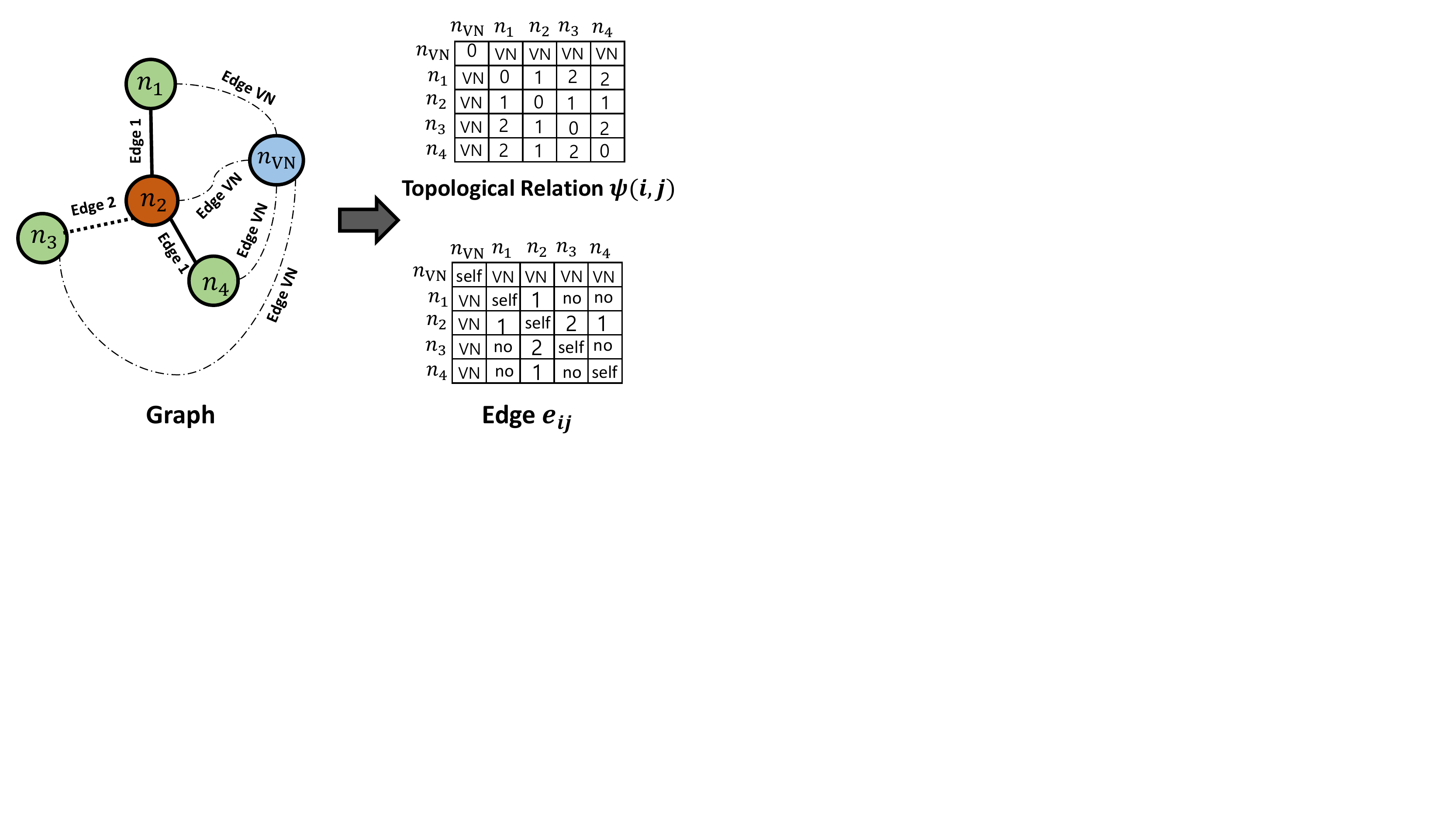}}
\vspace{-5cm}
\caption{
% Illustration of the proposed Graph Relative Positional Encoding.
% (a) shows an example of how GRPE process relative relation between nodes.
% In the example, we set the $L$ to 2.
% (b) describes our self-attention mechanism.
% Our two relative positional encodings, topology encoding and edge encoding, are used to encode graph on both attention map and value.
Illustration of how GRPE processes relative relation between nodes.
We add virtual node $n_{\text{VN}}$ to represent entire graph. 
For topological relations, we adopt the shortest path distance between nodes.
The virtual node is connected with all other nodes with a special edge VN and a special topological relation VN.
}
\label{fig:figure3}
\end{figure}

% We reformulate relative positional encoding for 1D sequence data \citep{shaw2018self} to incorporate graph-specific properties, such as topology relation or edges between nodes.
% We reformulate relative positional for 1D sequence data \citep{shaw2018self} to incorporate graph specific properties, such as topology relation or edges between nodes.
Our distinction is twofold.
First, we integrate interaction between node and graph structural information on the attention map.
For that, we propose node-aware attention which considers the interactions existing in two pairs: node-topology relation and node-edge relation.
Second, we also encode the topological information of a graph to the hidden representation of self-attention. For that, we propose graph-encoded values that directly encode relative positional information on the features of value by addition. 
Our node-aware attention applies the attention mechanism in a node-wise manner, while our graph-encoded value applies the attention mechanism in a channel-wise manner.

\subsection{Relative Positional Encoding for Graph}

We define two encodings to represent relative positional relation between two nodes in a graph.
The first is topology encoding $\mathcal{P}$, and we define the encodings for query, key, and value respectively: $\mathcal{P}^\text{query}, \mathcal{P}^\text{key}, \mathcal{P}^\text{value}\in\mathbb{R}^{L\times{d_{z}}}$.
Each vector of $\mathcal{P}$ represents the topological relation between two nodes, \eg, $\mathcal{P}_l$ represents the topological relation of two nodes where their shortest path distance is $l$. 
%$d_{z}$ is the dimension size of hidden features and $L$ is the maximum shortest path distance that our method considers.
$L$ is the maximum shortest path distance that our method considers.

The second is edge encoding $\mathcal{E}$, and we define the encodings for query, key, and value respectively: $\mathcal{E}^\text{query},\mathcal{E}^\text{key},\mathcal{E}^\text{value}\in\mathbb{R}^{E\times{d_{z}}}$. % \woonggi{where $E$ is the number of edge types}.
$\mathcal{E}_{e_{ij}}$ is a vector representing edge between two nodes $n_i$ and $n_j$.
$E$ is the number of types of edge.
The topology encodings and the edge encodings are shared throughout all layers.

\subsection{Node-Aware Attention}
\label{sec:context_aware}

% We propose two terms to encode graph on the attention map with two newly proposed encodings.
We propose two attention maps to encode a graph on self-attention.
%while considering interaction between node features and relative relation in graph, \eg, topology relation or edge.
The first attention map is $a^{\text{topology}}$.
It encodes graph by considering interaction between node feature and topological relation of graph.

\begin{equation}
\label{eq:ours_query_contextualized}
a^{\text{topology}}_{ij} = q_i{\cdot}\mathcal{P}^{\text{{query}}}_{\psi(i,j)}+k_j{\cdot}{\mathcal{P}^\text{{key}}_{\psi(i,j)}}
\end{equation}

%Likewise, the second term $b^{\text{key}}$ is a key contextualized bias that considers the interaction between the key feature and the graph information.
The second attention map is $a^{\text{edge}}$.
%It encodes graph by considering interaction between node feature of key and relative relation in graph.
It encodes graph by considering interaction between node feature and edge in graph.
%is a key contextualized bias that considers the interaction between the key feature and the graph information.
% Likewise, key contextualized bias considers the topology relation and edge types between two nodes with respect to the features of key and graph structure embeddings of query.
\begin{equation}
\label{eq:ours_key_contextualized}
a^{\text{edge}}_{ij} = q_i{\cdot}\mathcal{E}^{\text{query}}_{e_{ij}}+k_j{\cdot}{\mathcal{E}^\text{key}_{e_{ij}}}
\end{equation}

Finally, the two attention maps are added to scaled dot product attention map to encode graph information.

\begin{equation}
\label{eq:ours_attention_map}
a_{ij} = \frac{q_i{\cdot}k_j + a^{\text{topology}}_{ij} + a^{\text{edge}}_{ij}}{\sqrt{d_z}}.
\end{equation}

Our two attention maps consider topology and edge type, as node-topology and node-edge relation.
Meanwhile, Graphormer did not consider the interaction with node feature, such that
% Our two terms considers the interaction between the node features and relative relation existing in a graph, 
% Our two biases are parameterized by the hidden features (query and key) and graph information (structural relationship and edge) using the dot-product operations.
% , \eg, $q$ to $\mathcal{P}^{\text{key}}$ or $k$ to $\mathcal{E}^{\text{query}}$.
 two nodes with the same distance apart have the same bias $b_{\psi(i,j)}$ on Eq~\ref{eq:graphormer_attention}.
In contrast, our $a^{\text{topology}}$ enables to deploy different values according to the node features of query and key. 
%Therefore, our method attains a more expressive attention map.

\subsection{Graph-Encoded Value}
\label{sec:graph_embedded_value}

Another distinction is that our method directly encodes graph information into the hidden features of value, as well.
Specifically, we encode a graph to the hidden features of self-attention, where values are weighted summed with the attention map, for both topology and edge encoding
 via summation:
\begin{align}
\label{eq:graph_structure_value}
z_i = \sum_{j=1}^{N}\hat{a}_{ij}(v_j + \mathcal{P}_{\psi(i,j)}^\text{value} + \mathcal{E}_{e_{ij}}^\text{value}).
\end{align}

While the attention weight $\hat{a}$ is applied equally for all channels (node-wise attention),  our graph-encoded value enriches 
the feature of each channel, and enables channel-wise attention as well.
%our proposed graph-encoded value is a channel-wise attention mechanism.
%Therefore, node-aware attention is a node-wise attention mechanism while 
This relative positional encoding enables to encode relative position directly on hidden features, without 
previous approaches that encode position on input $x$ with a linearized graph, as in Figure 1a, \eg, centrality encoding~\citep{ying2021transformers} or graph Laplacian~\citep{san_kreuzer2021rethinking,gt_dwivedi2020generalization},
where linearization sacrifices the preciseness of position.
% Unlike previous approaches where graph is encoded on the initial input $x$, \eg, centrality encoding~\citep{ying2021transformers} or graph Laplacian~\citep{san_kreuzer2021rethinking,gt_dwivedi2020generalization}

\subsection{Complexity Analysis}
\label{sec:efficient_impl}

A naive implementation of computing all pairs of $a^{\text{topology}}$ requires time complexity of $O({N^2}d_z)$, since it requires performing the dot product between all node pairs.
Instead, we pre-compute the dot product of all possible pairs of node features and topology encoding vectors $\mathcal{P}$ which requires time complexity of $O(NL{d_z})$. 
Then we assign the pre-computed value according to the indices of node pairs.
Likewise, for the $a^{\text{edge}}$, we pre-compute the dot product of all possible pairs of node features and edge encoding vectors $\mathcal{E}^{\text{query}}$ which requires time complexity of $O(NE{d_z})$.
The time complexity is reduced significantly with our implementation since $L$ and $E$ are much smaller than the number of nodes $N$.
% We apply the same implementation  too.

\section{Experiment}
\label{sec:experiment}

\begin{table*}
\centering
\caption {Configurations of models that we utilize throughout our experiments.} 
\label{tab:models}

\scalebox{1.0}{
\begin{tabular}{l|c|ccccc}
\toprule
\multicolumn{1}{c|}{Model Configurations} & \# Params & \# Layers & Hidden dim [$d_{z}$] & FFN layer dim & \# Heads \\ \hline
% GSA-Tiny & 106k & 4  & 64 & 64  & 8    \\
% GSA-Tiny2x & 409k & 4 & 128 & 128 & 16 \\
GRPE-Tiny & 106k & 4  & 64 & 64  & 8    \\
GRPE-Small  & 489k & 12 & 80 & 80 & 8  \\
GRPE-Standard & 46.2M & 12 & 768 & 768 & 32 \\
GRPE-Large  & 118.3M  & 18  & 1024 & 1024 & 32 \\ 
\bottomrule
\end{tabular}}
\end{table*}

\begin{table}
\centering
\caption {Results on ZINC. $^*$ indicates a fine-tuned model. The lower the better.} 
\label{tab:zinc}
\vspace{-0.2cm}
	\scalebox{0.8}{
\begin{tabular}{ccc}
\toprule
\multicolumn{1}{c|}{Method} & \multicolumn{1}{c|}{\#Params} & MAE \\ \hline
\multicolumn{1}{l|}{GIN \citep{gin_xu2018powerful}}  & \multicolumn{1}{c|}{510k} & 
\multicolumn{1}{c}{0.526$\pm$0.051}\\
% \multicolumn{1}{l|}{GraphSage~\citep{graphsage_hamilton2017inductive}}  & \multicolumn{1}{c|}{505k} & \multicolumn{1}{c}{0.398$\pm$0.002}\\
\multicolumn{1}{l|}{GAT~\citep{velivckovic2017graph}}  & \multicolumn{1}{c|}{531k} & 
\multicolumn{1}{c}{0.384$\pm$0.007} \\
\multicolumn{1}{l|}{GCN~\citep{gcn_kipf2016semi}}  & \multicolumn{1}{c|}{505k} & 
\multicolumn{1}{c}{0.367$\pm$0.011} \\
\multicolumn{1}{l|}{GatedGCN~\citep{gated_bresson2017residual}}  & \multicolumn{1}{c|}{505k} & 
\multicolumn{1}{c}{0.214$\pm$0.006} \\
\multicolumn{1}{l|}{MPNN$^*$ (sum)~\citep{mpnn_gilmer2017neural}}  & \multicolumn{1}{c|}{481k} & \multicolumn{1}{c}{0.145$\pm$0.007} \\
\multicolumn{1}{l|}{PNA~\citep{pna_corso2020principal}}  & \multicolumn{1}{c|}{387k} & 
\multicolumn{1}{c}{0.142$\pm$0.010} \\
\hline
\multicolumn{1}{l|}{$\dagger$GT~\citep{gt_dwivedi2020generalization}}  & \multicolumn{1}{c|}{589k} & 
\multicolumn{1}{c}{0.226$\pm$0.014} \\
\multicolumn{1}{l|}{$\dagger$SAN~\citep{san_kreuzer2021rethinking}}  & \multicolumn{1}{c|}{509k} & 
\multicolumn{1}{c}{0.139$\pm$0.006} \\
\multicolumn{1}{l|}{$\dagger$Graphormer (slim) \citep{ying2021transformers}} & \multicolumn{1}{c|}{489k} & 
\multicolumn{1}{c}{0.122$\pm$0.006} \\
\multicolumn{1}{l|}{$\dagger$EGT \cite{hussain2021edge}} &\multicolumn{1}{c|}{$\approx$500k} & 
\multicolumn{1}{c}{0.108$\pm$0.009} \\    
\hline
\multicolumn{1}{l|}{$\dagger$GRPE-Small (Ours)} &\multicolumn{1}{c|}{489k} & 
\multicolumn{1}{c}{ \bf{0.094$\pm$0.002}} \\    
\bottomrule
\end{tabular}}
\end{table}

\begin{table}
\centering
\caption {Results on MolHIV. $^*$ indicates a fine-tuned model. The higher the better.} 
\label{tab:hiv}
\vspace{-0.2cm}
\hspace{-0.5cm}
\scalebox{0.8}{
\begin{tabular}{ccc}
\toprule
\multicolumn{1}{c|}{Method} & \multicolumn{1}{c|}{\#Params} & AUC (\%) \\ \hline
\multicolumn{1}{l|}{GCN-GraphNorm~\citep{cai2021graphnorm}}  & \multicolumn{1}{c|}{526k} & 
\multicolumn{1}{c}{78.83$\pm$1.00}\\
\multicolumn{1}{l|}{PNA~\citep{pna_corso2020principal}}  & \multicolumn{1}{c|}{326k} & 
\multicolumn{1}{c}{79.05$\pm$ 1.32}\\
\multicolumn{1}{l|}{PHC-GNN~\citep{phc_le2021parameterized}}  & \multicolumn{1}{c|}{111k} & 
\multicolumn{1}{c}{79.34$\pm$ 1.16} \\
\multicolumn{1}{l|}{DeeperGCN-FLAG~\citep{deepergn_li2020deepergcn}}  & \multicolumn{1}{c|}{532k} & 
\multicolumn{1}{c}{79.42$\pm$ 1.20} \\
\multicolumn{1}{l|}{DGN~\citep{dgn_beani2021directional}}  & \multicolumn{1}{c|}{114k} & 
\multicolumn{1}{c}{79.70$\pm$ 0.97} \\
% \multicolumn{1}{l|}{GIN-VN$^*$~\citep{gin_xu2018powerful}}  & \multicolumn{1}{c|}{3.3M} & 
% \multicolumn{1}{c}{77.80$\pm$ 1.82} \\
\multicolumn{1}{l|}{GIN$^*$~\citep{gin_xu2018powerful}}  & \multicolumn{1}{c|}{3.3M} & 
\multicolumn{1}{c}{77.80$\pm$ 1.82} \\
\hline
\multicolumn{1}{l|}{$\dagger$Graphormer-FLAG$^*$~\cite{ying2021transformers}}  & \multicolumn{1}{c|}{47.0M} &  \multicolumn{1}{c}{80.51$\pm$0.53} \\
\multicolumn{1}{l|}{$\dagger$EGT-Large$^*$ ~\cite{hussain2021edge}} & \multicolumn{1}{c|}{110.8M} &  \multicolumn{1}{c}{80.60$\pm$0.65} \\
\hline
\multicolumn{1}{l|}{$\dagger$GRPE-Standard$^*$ (Ours)} &\multicolumn{1}{c|}{46.2M} & 
\multicolumn{1}{c}{ \bf{81.39  $\pm$ 0.49}} \\    
\bottomrule
\end{tabular}}
\end{table}

\begin{table}
\centering
\caption {Results on MolPCBA.  $^*$ indicates a fine-tuned model. The higher the better.} 
\label{tab:pcba}
\scalebox{0.8}{
\begin{tabular}{ccc}
\toprule
\multicolumn{1}{c|}{Method} & \multicolumn{1}{c|}{\#Params} & AP (\%) \\ \hline
\multicolumn{1}{l|}{DeeperGCN+FLAG~\citep{deepergn_li2020deepergcn}}  & \multicolumn{1}{c|}{5.6M} & 
\multicolumn{1}{c}{28.42$\pm$ 0.43}\\
\multicolumn{1}{l|}{DGN~\citep{dgn_beani2021directional}}  & \multicolumn{1}{c|}{6.7M} & 
\multicolumn{1}{c}{28.85$\pm$ 0.30}\\
% \multicolumn{1}{l|}{GINE-VN~\citep{gine_brossard2020graph}}  & \multicolumn{1}{c|}{6.1M} & 
% \multicolumn{1}{c}{29.17$\pm$ 0.15} \\
\multicolumn{1}{l|}{PHC-GNN~\citep{phc_le2021parameterized}}  & \multicolumn{1}{c|}{1.7M} & 
\multicolumn{1}{c}{29.47$\pm$ 0.26} \\
\multicolumn{1}{l|}{GINE~\citep{gine_brossard2020graph}}  & \multicolumn{1}{c|}{6.1M} & 
\multicolumn{1}{c}{29.79$\pm$ 0.30} \\
\multicolumn{1}{l|}{GIN$^*$~\citep{gin_xu2018powerful} }  & \multicolumn{1}{c|}{3.4M} & 
\multicolumn{1}{c}{29.02$\pm$ 0.17} \\
\hline
\multicolumn{1}{l|}{$\dagger$Graphormer-FLAG$^*$~\citep{ying2021transformers}}  & \multicolumn{1}{c|}{119.5M} & 
\multicolumn{1}{c}{31.39$\pm$ 0.32} \\
\multicolumn{1}{l|}{$\dagger$EGT-Large$^*$~\citep{hussain2021edge}}  & \multicolumn{1}{c|}{110.8M} & 
\multicolumn{1}{c}{29.61$\pm$0.24} \\
\hline
\multicolumn{1}{l|}{$\dagger$GRPE-Standard$^*$ (Ours)} &\multicolumn{1}{c|}{ 46.2M} & 
\multicolumn{1}{c}{30.77$\pm$ 0.07} \\   
\multicolumn{1}{l|}{$\dagger$GRPE-Large$^*$ (Ours)} &\multicolumn{1}{c|}{ 118.3M} & 
\multicolumn{1}{c}{\bf{31.50}$\pm$ 0.10} \\  
\bottomrule
\end{tabular}
}
\end{table}

\begin{table}
\centering
\caption {Results on PATTERN. The higher the better.} 
% \caption {Results on PATTERN. The higher the better. For models with 100k and 500k parameters, we adopt GRPE-Tiny and GRPE-Small respectively.} 
\label{tab:pattern}
\vspace{-0.2cm}
\hspace{-0.6cm}
	\scalebox{0.8}{
\begin{tabular}{ccc}
\toprule
\multicolumn{1}{c|}{Method} & \multicolumn{2}{c}{Weighted Accuracy} \\ 
\multicolumn{1}{c|}{} & \multicolumn{1}{c}{\#Params} & \multicolumn{1}{c}{\#Params} \\ 
\multicolumn{1}{c|}{} & \multicolumn{1}{c}{$\approx$100k} & \multicolumn{1}{c}{$\approx$500k} \\ \hline

\multicolumn{1}{l|}{GIN \citep{gin_xu2018powerful}}  & \multicolumn{1}{c}{85.590$\pm$0.011} & 
\multicolumn{1}{c}{85.387$\pm$0.136}\\
% \multicolumn{1}{l|}{GraphSage~\citep{graphsage_hamilton2017inductive}}  & \multicolumn{1}{c}{50.516$\pm$0.001} & \multicolumn{1}{c}{50.492 $\pm$0.001}\\
\multicolumn{1}{l|}{GAT~\citep{velivckovic2017graph}}  & \multicolumn{1}{c}{75.824$\pm$1.823} & 
\multicolumn{1}{c}{78.271$\pm$0.186} \\
\multicolumn{1}{l|}{GCN~\citep{gcn_kipf2016semi}}  & \multicolumn{1}{c}{63.880$\pm$0.074} & 
\multicolumn{1}{c}{71.892$\pm$0.334} \\
\multicolumn{1}{l|}{GatedGCN~\citep{gated_bresson2017residual}}  & \multicolumn{1}{c}{84.480$\pm$0.122} & 
\multicolumn{1}{c}{86.508$\pm$0.085} \\
% \multicolumn{1}{l|}{MPNN$^*$ (sum)~\citep{mpnn_gilmer2017neural}}  & \multicolumn{1}{c}{481k} & \multicolumn{1}{c}{0.145$\pm$0.007} \\
\multicolumn{1}{l|}{PNA~\citep{pna_corso2020principal}}  & \multicolumn{1}{c}{86.567$\pm$0.075} & 
\multicolumn{1}{c}{-} \\
\hline
\multicolumn{1}{l|}{$\dagger$GT~\citep{gt_dwivedi2020generalization}}  & \multicolumn{1}{c}{-} & 
\multicolumn{1}{c}{84.808$\pm$0.068} \\
\multicolumn{1}{l|}{$\dagger$SAN~\citep{san_kreuzer2021rethinking}}  & \multicolumn{1}{c}{-} & 
\multicolumn{1}{c}{86.581$\pm$0.037} \\
\multicolumn{1}{l|}{$\dagger$Graphormer (slim) \citep{ying2021transformers}} & \multicolumn{1}{c}{-} & 
\multicolumn{1}{c}{86.650$\pm$0.033} \\
\multicolumn{1}{l|}{$\dagger$EGT \citep{hussain2021edge}} & \multicolumn{1}{c}{\textbf{86.816$\pm$0.027}} & 
\multicolumn{1}{c}{86.821$\pm$0.020} \\
\hline
\multicolumn{1}{l|}{$\dagger$GRPE (Ours)} &\multicolumn{1}{c}{83.105$\pm$0.045} & 
\multicolumn{1}{c}{\textbf{87.020$\pm$0.042}} \\    
\bottomrule
\end{tabular}}
\end{table}

\begin{table}[t]
\centering
\caption {Results on CLUSTER. The number of parameters is around 500K for all models. The higher the better.}
\label{tab:cluster}
\vspace{-0.2cm}
	\scalebox{0.88}{
\begin{tabular}{cc}
\toprule
\multicolumn{1}{c|}{Method} & \multicolumn{1}{c}{Weighted Accuracy} \\ \hline
% \multicolumn{1}{c|}{} & \multicolumn{1}{c}{\#Params $ \approx $500K} \\ \hline

\multicolumn{1}{l|}{GIN \citep{gin_xu2018powerful}}  & \multicolumn{1}{c}{64.716$\pm$1.553}\\
% \multicolumn{1}{l|}{GraphSage~\citep{graphsage_hamilton2017inductive}}  & \multicolumn{1}{c}{50.516$\pm$0.001} & \multicolumn{1}{c}{50.492 $\pm$0.001}\\
\multicolumn{1}{l|}{GAT~\citep{velivckovic2017graph}}  & \multicolumn{1}{c}{70.587$\pm$0.447} \\
\multicolumn{1}{l|}{GCN~\citep{gcn_kipf2016semi}}  & \multicolumn{1}{c}{68.498$\pm$0.976}  \\
\multicolumn{1}{l|}{GatedGCN~\citep{gated_bresson2017residual}}  & \multicolumn{1}{c}{76.082$\pm$0.196} \\
\hline
% \multicolumn{1}{l|}{MPNN$^*$ (sum)~\citep{mpnn_gilmer2017neural}}  & \multicolumn{1}{c}{481k} & \multicolumn{1}{c}{0.145$\pm$0.007} \\
% \multicolumn{1}{l|}{PNA~\citep{pna_corso2020principal}}  & \multicolumn{1}{c}{86.567$\pm$0.075}  \\
% \hline
\multicolumn{1}{l|}{$\dagger$GT~\citep{gt_dwivedi2020generalization}}  & \multicolumn{1}{c}{73.169$\pm$0.622}  \\
\multicolumn{1}{l|}{$\dagger$SAN~\citep{san_kreuzer2021rethinking}}  & \multicolumn{1}{c}{76.691$\pm$0.650} \\
\multicolumn{1}{l|}{$\dagger$Graphormer (slim) \citep{ying2021transformers}} & \multicolumn{1}{c}{74.660$\pm$0.236}  \\
\multicolumn{1}{l|}{$\dagger$EGT \citep{hussain2021edge}} & \multicolumn{1}{c}{79.232$\pm$0.348}  \\
\hline
\multicolumn{1}{l|}{$\dagger$GRPE-Small (Ours)} &\multicolumn{1}{c}{\textbf{81.586$\pm$0.190}} \\    
\bottomrule
\end{tabular}}
\end{table}

\begin{table*}[ht!]
\centering
\caption {Results on PCQM4M. * indicates the results are from the official leaderboard. VN indicates that the model used virtual node. The lower the better.} 
\label{tab:pcqm4m-lsc}
\vspace{0.1cm}
	\scalebox{1.0}{
\begin{tabular}{cccccc}
\toprule
\multicolumn{1}{c|}{Method} & \multicolumn{1}{c|}{\#Params} & Train MAE & Validate MAE & Test MAE \\ \hline
\multicolumn{1}{l|}{GCN~\citep{gcn_kipf2016semi} }  & \multicolumn{1}{c|}{2.0M} & 
\multicolumn{1}{c}{0.1318} & \multicolumn{1}{c}{0.1691 (0.1684$^*$)} & \multicolumn{1}{c}{0.1838$^*$} &  \\
\multicolumn{1}{l|}{GIN~\citep{gin_xu2018powerful} }  & \multicolumn{1}{c|}{3.8M} & 
\multicolumn{1}{c}{0.1203} & \multicolumn{1}{c}{0.1537 (0.1536$^*$)} & \multicolumn{1}{c}{0.1678$^*$} &  \\
\multicolumn{1}{l|}{GCN-VN~\citep{gcn_kipf2016semi}} & \multicolumn{1}{c|}{4.9M} & 
\multicolumn{1}{c}{0.1225} & \multicolumn{1}{c}{0.1485 (0.1510$^*$)} & \multicolumn{1}{c}{0.1579$^*$} &  \\
\multicolumn{1}{l|}{GIN-VN~\citep{gin_xu2018powerful}}  & \multicolumn{1}{c|}{6.7M} & 
\multicolumn{1}{c}{0.1150} & \multicolumn{1}{c}{0.1395 (0.1396$^*$)} & \multicolumn{1}{c}{0.1487$^*$} & \\
\multicolumn{1}{l|}{GINE-VN~\citep{gine_brossard2020graph}} & \multicolumn{1}{c|}{13.2M} & 
\multicolumn{1}{c}{0.1248} & \multicolumn{1}{c}{0.1430} & \multicolumn{1}{c}{-} &  \\
\multicolumn{1}{l|}{DeeperGCN-VN~\citep{deepergn_li2020deepergcn}} & \multicolumn{1}{c|}{25.5M} &
\multicolumn{1}{c}{0.1059} & \multicolumn{1}{c}{0.1398} & \multicolumn{1}{c}{-} &  \\
\hline
\multicolumn{1}{l|}{$\dagger$GT~\citep{gt_dwivedi2020generalization}} & \multicolumn{1}{c|}{0.6M} &
\multicolumn{1}{c}{0.0944} & \multicolumn{1}{c}{0.1400} & \multicolumn{1}{c}{-} &  \\  
\multicolumn{1}{l|}{$\dagger$GT-wide~\citep{gt_dwivedi2020generalization}} & \multicolumn{1}{c|}{83.2M} &
\multicolumn{1}{c}{0.0955} & \multicolumn{1}{c}{0.1408} & \multicolumn{1}{c}{-} &  \\  
\multicolumn{1}{l|}{$\dagger$Graphormer (small)~\citep{ying2021transformers}} & \multicolumn{1}{c|}{12.5M} &
\multicolumn{1}{c}{0.0778} & \multicolumn{1}{c}{0.1264} & \multicolumn{1}{c}{-} &  \\  
\multicolumn{1}{l|}{$\dagger$Graphormer~\citep{ying2021transformers}} & \multicolumn{1}{c|}{47.1M} &
\multicolumn{1}{c}{0.0582} & \multicolumn{1}{c}{0.1234} & \multicolumn{1}{c}{\textbf{0.1328}} &  \\    
\multicolumn{1}{l|}{$\dagger$EGT-Medium~\citep{hussain2021edge}} & \multicolumn{1}{c|}{47.4M} &
\multicolumn{1}{c}{-} & \multicolumn{1}{c}{\bf{0.1224}} & \multicolumn{1}{c}{-} &  \\    
\hline
\multicolumn{1}{l|}{$\dagger$GRPE-Standard (Ours)} &\multicolumn{1}{c|}{46.2M} & 
\multicolumn{1}{c}{\bf{0.0349}} & \multicolumn{1}{c}{0.1225} & \multicolumn{1}{c}{-} &  \\    

\bottomrule
\end{tabular}}
\end{table*}

\begin{table*}[!ht]
\centering
\caption {Results on PCQM4Mv2. VN indicates that the model used the virtual node. The lower the better.} 
\label{tab:pcqm4m-lsc-v2}
\vspace{0.1cm}
	\scalebox{1.0}{
\begin{tabular}{ccccc}
\toprule
\multicolumn{1}{c|}{Method} & \multicolumn{1}{c|}{\#Params} & Validate MAE & Test-dev MAE \\ \hline
\multicolumn{1}{l|}{GCN~\citep{gcn_kipf2016semi} }  & \multicolumn{1}{c|}{2.0M} & \multicolumn{1}{c}{0.1379} & \multicolumn{1}{c}{0.1398} &  \\
\multicolumn{1}{l|}{GIN~\citep{gin_xu2018powerful} }  & \multicolumn{1}{c|}{3.8M} & \multicolumn{1}{c}{0.1195} & \multicolumn{1}{c}{0.1218} &  \\
\multicolumn{1}{l|}{GCN-VN~\citep{gcn_kipf2016semi}} & \multicolumn{1}{c|}{4.9M} & \multicolumn{1}{c}{0.1153} & \multicolumn{1}{c}{0.1152} &  \\
\multicolumn{1}{l|}{GIN-VN~\citep{gin_xu2018powerful}}  & \multicolumn{1}{c|}{6.7M} & \multicolumn{1}{c}{0.1083} & \multicolumn{1}{c}{0.1084} & \\
\hline
\multicolumn{1}{l|}{$\dagger$EGT-Medium~\citep{hussain2021edge}}  & \multicolumn{1}{c|}{47.4M} & \multicolumn{1}{c}{0.0881} & \multicolumn{1}{c}{-} & \\
\multicolumn{1}{l|}{$\dagger$EGT-Large~\citep{hussain2021edge}}  & \multicolumn{1}{c|}{89.3M} & \multicolumn{1}{c}{0.0869} & \multicolumn{1}{c}{\textbf{0.0872}} & \\
% \multicolumn{1}{l|}{GINE-VN~\cite{gine_brossard2020graph}} & \multicolumn{1}{c|}{13.2M} & \multicolumn{1}{c}{0.1430} & \multicolumn{1}{c}{-} &  \\
% \multicolumn{1}{l|}{MLP-Fingerprint~\cite{deepergn_li2020deepergcn}} & \multicolumn{1}{c|}{25.5M} & \multicolumn{1}{c}{0.1398} & \multicolumn{1}{c}{-} &  \\
\hline
\multicolumn{1}{l|}{$\dagger$GRPE-Standard (Ours) } &\multicolumn{1}{c|}{46.2M} &  \multicolumn{1}{c}{0.0890} & \multicolumn{1}{c}{0.0898} &  \\    
\multicolumn{1}{l|}{$\dagger$GRPE-Large (Ours) } &\multicolumn{1}{c|}{118.3M} &  \multicolumn{1}{c}{\textbf{0.0866}} & \multicolumn{1}{c}{-} &  \\    
\bottomrule
\end{tabular}}
\end{table*}

\subsection{Implementation Details}

\subsubsection{Virtual node}

Following \citeauthor{mpnn_gilmer2017neural} and \citeauthor{ying2021transformers}, we adopt a special node called virtual node which is connected to all other nodes.
The role of a virtual node is similar to special tokens such as a classification token \citep{devlin2018bert}, where its output feature $z$ is used as the input for the branch that predicts downstream tasks.
We additionally define two encoding vectors to define both topology relation $\mathcal{P}_{\text{VN}}$ and edge $\mathcal{E}_{\text{VN}}$ for query, key and value respectively.
Note that, the virtual node does not involve to find the shortest path between two nodes.
Throughout all experiments, we add the virtual node to a graph to perform the downstream tasks.
Figure~\ref{fig:figure3} illustrates how a virtual node is connected with other nodes.

\subsubsection{Topological relation}
We utilize shortest path distance $\psi(i, j)$ to describe the topological relation between two nodes $n_i$ and $n_j$.
$L$  is the maximum distance of the shortest path that we consider, and we utilize a special encoding vector $\mathcal{P}_{\text{far}}$ for the node pairs with a distance more than $L$.
For the nodes pairs that are unreachable, we utilize another special encoding vector $\mathcal{P}_{\text{unreachable}}$.
Finally, for the pairs that are connected with the virtual node, we utilize another special encoding vector $\mathcal{P}_{\text{VN}}$.
% The structure embedding $\mathcal{P}_K$ is a vector to represents relationship of two nodes of $K$-hop apart. 
% Therefore, there are $L+3$ embedding vectors in each $\mathcal{P}$.
Figure~\ref{fig:figure3} illustrates how topological relation is processed.

\subsubsection{Edge}

Some pair of nodes are not connected with edges.
Therefore, we utilize a special encoding vector $\mathcal{E}_{\text{no}}$ for the pairs of node that are not connected with any edges; $\{(n_i,n_j) | i\neq{j}\,\text{and} \,j \notin \mathcal{N}_i\}_{i=1:|N|}$.
For the pair of two identical nodes where $i = j$, we use a special embedding vector $\mathcal{E}_{\text{self}}$.
Finally, for the pairs that are connected with the virtual node, we utilize another special encoding vector $\mathcal{E}_{\text{VN}}$.
Figure~\ref{fig:figure3} illustrates how edges are processed.

\subsection{Graph Classification and Regression}

 We summarize the model configurations of our experiments in Table~\ref{tab:models}.
 For all results, 
$\dagger$ indicates the models adopted Transformer for learning graph representation
and text in \textbf{bold}
indicates the best result.
% We choose the hyper parameters of the models for fair comparison with other methods regarding the number of parameters.

% We validate our method on the task of the graph-level property prediction with benchmarking datasets with MIT license: OGBG-MolPCBA (MolPCBA) \cite{hu2020open}, OGBG-MolHIV (MolHIV) \cite{hu2020open} and ZINC from the benchmarking GNN \cite{dwivedi2020benchmarking}.
% MolHIV is a small dataset that consists of 41,127 graphs. 
We first validate our method on the tasks of the molecule property prediction such as OGBG-MolPCBA (MolPCBA) \citep{hu2020open}, OGBG-MolHIV (MolHIV) \citep{hu2020open} and ZINC \citep{dwivedi2020benchmarking}.
MolPCBA consists of 437,929 graphs and the task is to predict multiple binary labels indicating various molecule properties. The evaluation metric is average precision (AP). 
MolHIV is a small dataset that consists of 41,127 graphs. 
The task is to predict a binary label indicating whether a molecule inhibits HIV virus replication or not. 
The evaluation metric is area under the curve (AUC).
ZINC is also a small dataset that consists of 12,000 graphs, and the task is to regress a molecule property. 
% Note that we follow the dataset split from the \citep{dwivedi2020benchmarking}.
The evaluation metric is mean absolute error (MAE). 
All experiments are conducted for 5 times, and we report the mean and the standard deviation of the experiments.

We adopt the linear learning rate decay, and the learning rate starts from $2{*}10^{-4}$ and ends at $1{*}10^{-9}$.
We set $L$ to 5. 
For ZINC dataset, we adopt GRPE-Small configuration with less than 500k parameters for a fair comparison.
% We apply dropout~\citep{srivastava2014dropout} on the attention map and the input to feed-forward neural networks with the probability of 0.1.
% For MolHIV dataset, we adopt GSA-Standard configuration for fair comparison with Graphormer~\citep{ying2021transformers}. 
% For MolPCBA dataset, we adopt GSA-Standard and GSA-Large configurations.
% We apply dropout on the attention map and the input to feed-forward neural networks with the probability of 0.3 
For MolHIV and MolPCBA datasets, we initialize the parameter of the models with the weight of a pretrained model trained on PCQM4M~\citep{hu2020open} dataset.

Table~\ref{tab:zinc} shows the results on ZINC dataset, our model achieve state-of-the-art MAE score.
% Our method achieves far better score than GCN-based methods, and achieves significantly better score than Transformer-based methods.
Table~\ref{tab:pcba} shows the results on MolPCBA dataset, our model achieves state-of-the-art AP score. 
Table~\ref{tab:hiv} shows the results on MolHIV dataset, our model achieves the state-of-the-art AUC with less parameters than Graphormer.

\subsection{Node Classification}
We validate our method on the task of the node-wise classification such as PATTERN and CLUSTER \cite{dwivedi2020benchmarking}. 
PATTERN consists of 14,000 graphs and the task is to recognize graph pattern where each node belong, and the number of classes are two.
CLUSTER consists of 12,000 graphs and the task is semi-supervised clustering where each node belong, and the number of classes are six.
The evaluation metric is the average node-level accuracy weighted with respect to the class sizes, we follow the evaluation code presented in the original work~\cite{dwivedi2020benchmarking}.

We adopt the linear learning rate decay, and the learning rate starts from $2{*}10^{-4}$ and ends at $1{*}10^{-9}$.
We set $L$ to 5. 
For both datasets, we adopt GRPE-Small with less than 500k parameters and adopt GRPE-Tiny with less than 100k parameters for a fair comparison.
Table~\ref{tab:pattern} shows the results on PATTERN, our models achieve state-of-the-art accuracy on 500k parameters.
Table~\ref{tab:cluster} shows the results on CLUSTER, our models achieve state-of-the-art weighted accuracy with a significant improvement.
% Our method achieves far better score than GCN-based methods, and achieves significantly better score than Transformer-based methods.
% Table~\ref{tab:cluster} shows the results on CLUSTER dataset, our model achieves state-of-the-art weighted accuracy.

\subsection{OGB Large Scale Challenge}

We validate our method on two datasets of OGB large scale challenge~\citep{hu2020open}.
The two datasets aim to predict the DFT-calculated HOMO-LUMO energy gap of molecules given their molecular graphs. 
We conduct experiments on both the PCQM4M and PCQM4Mv2 datasets, which are currently the biggest molecule property prediction datasets containing about 4 million graphs in total.
PCQM4Mv2 contains the DFT-calcuated 3D strcuture of molecules.
For our experiments, we only utilize 2D molecular graphs not 3D structures.
%We achieve the-state-of-the-art MAE on the validation set of both datasets.
Throughout experiments, we set $L$ to 5. 
We adopt a GRPE-Standard for fair comparisons with Graphormer.
% We dropout the attention map and the input to feed-forward neural networks with the probability of 0.1.
We linearly increase learning rate up to $2{*}10^{-4}$ for 3 epochs and linearly decay learning rate upto $1{*}10^{-9}$ for 400 epochs.
We are unable to measure the test MAE of PCQM4M, because the test dataset is deprecated as PCQM4Mv2 is newly released.

Table~\ref{tab:pcqm4m-lsc} shows the results on PCQM4M dataset. 
% Our model obtain the best train MAE showing that our model successfully memorizes training datasets with less parameters than Graphormer~\citep{ying2021transformers} or GT~\citep{gt_dwivedi2020generalization}.
Our model achieves the second best validation MAE score, but with a very small gap with the best model of about $0.0001$. 
Table.~\ref{tab:pcqm4m-lsc-v2} shows the results on PCQM4Mv2 dataset.
% There are not many methods tested on PCQM4Mv2, because the dataset is recently updated.
Our large model achieves the best result on the validation dataset.
We couldn't report the large model's test-dev MAE, since the evaluation server only allow one submission per week. 
We will make sure to report the result for the final draft.
For the models with a similar number of parameters GRPE-Standard and EGT-Medium, we achieved competitive results.
% However, our model use only half of the parameters compared to the best models.
% Note that, methods with the lower validation MAE obtain the better test MAE on both datasets.
% therefore we believe that our method will obtain the best MAE on test dataset once we are able to submit our results after publication.

% \newpage

% \subsection{Analysis}

\begin{figure*}
\centering
\hspace*{-3cm}\includegraphics[width=1.3\linewidth]{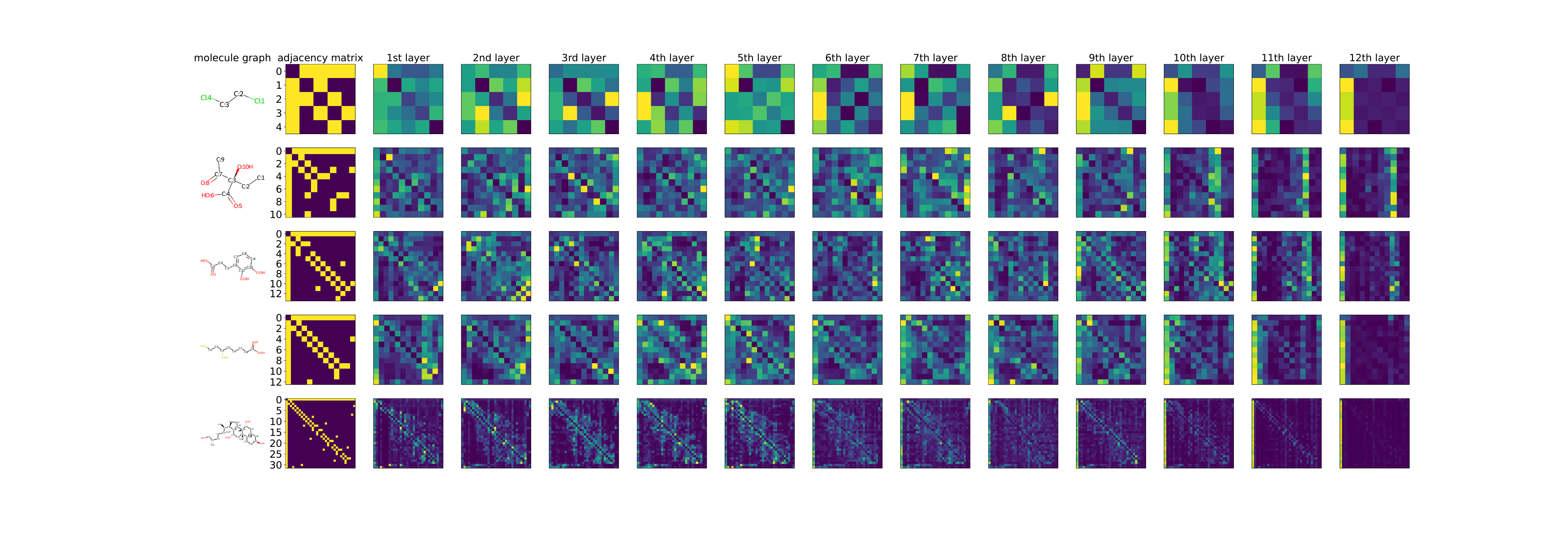}
\vspace{-1.3cm}
\caption{Visualization of adjacency matrix and average attention map (starting from third column) of each layer on molecule graph with various size. We added virtual node as the first node marked with zero index. For adjacency matrix, yellow and dark purple indicate the two nodes are connected and disconnected respectively. For attention map, brighter color (yellow) indicates higher attention and dark color indicates lower attention.}
\label{fig:viz_att}
\end{figure*}

\subsection{Ablation Study on Components of GRPE}
\label{sec:gsa_effect}

We validate the effect of components of GRPE on ZINC dataset.
We adopt GRPE-Small.
Table~\ref{tab:zinc_comp_ablation} shows the ablation study results.
The first row is identical to the plain Transformer without any positional encodings, and it obviously shows the highest error.
Adding either $a^{\text{topology}}$ or $a^{\text{edge}}$ lowers the error, and using them together further lowers the error.
Finally, adding our Graph-Encoded Value does help to improve the performance.

\begin{table}
\centering
\caption {Effects of components of GRPE on ZINC dataset. The lower the better.} 
\label{tab:zinc_comp_ablation}
\vspace{-0.2cm}
\begin{tabular}{cccc}
\toprule                   
% \multicolumn{2}{c}{Node-Aware Relative Self-Attention} & \multicolumn{1}{c}{\multirow{2}{*}{Graph-Encoded Value}} & \multicolumn{1}{c}{\multirow{2}{*}{Test MAE}}\\  
% \cline{1-2} \\
% \multicolumn{1}{c}{$b^{\text{spatial}}$} & \multicolumn{1}{c}{$b^{\text{edge}}$}
% \\ \hline

$a^{\text{topology}}$ & $a^{\text{edge}}$ & Graph-Encoded Value & Test MAE \\
\hline

- & - & - & 0.668 \\
$\checkmark$ & - & - & 0.267\\
- & $\checkmark$ & - & 0.218 \\
- & - & $\checkmark$ & 0.116 \\
$\checkmark$ & $\checkmark$ & - & 0.147 \\
$\checkmark$ & $\checkmark$ & $\checkmark$ & \textbf{0.093} \\

\bottomrule
\end{tabular}
\end{table}

\subsection{Effects of Maximum Shortest Path Distance $L$}

\begin{figure}
\begin{minipage}{0.49\textwidth}
\centering
\includegraphics[width=0.75\textwidth]{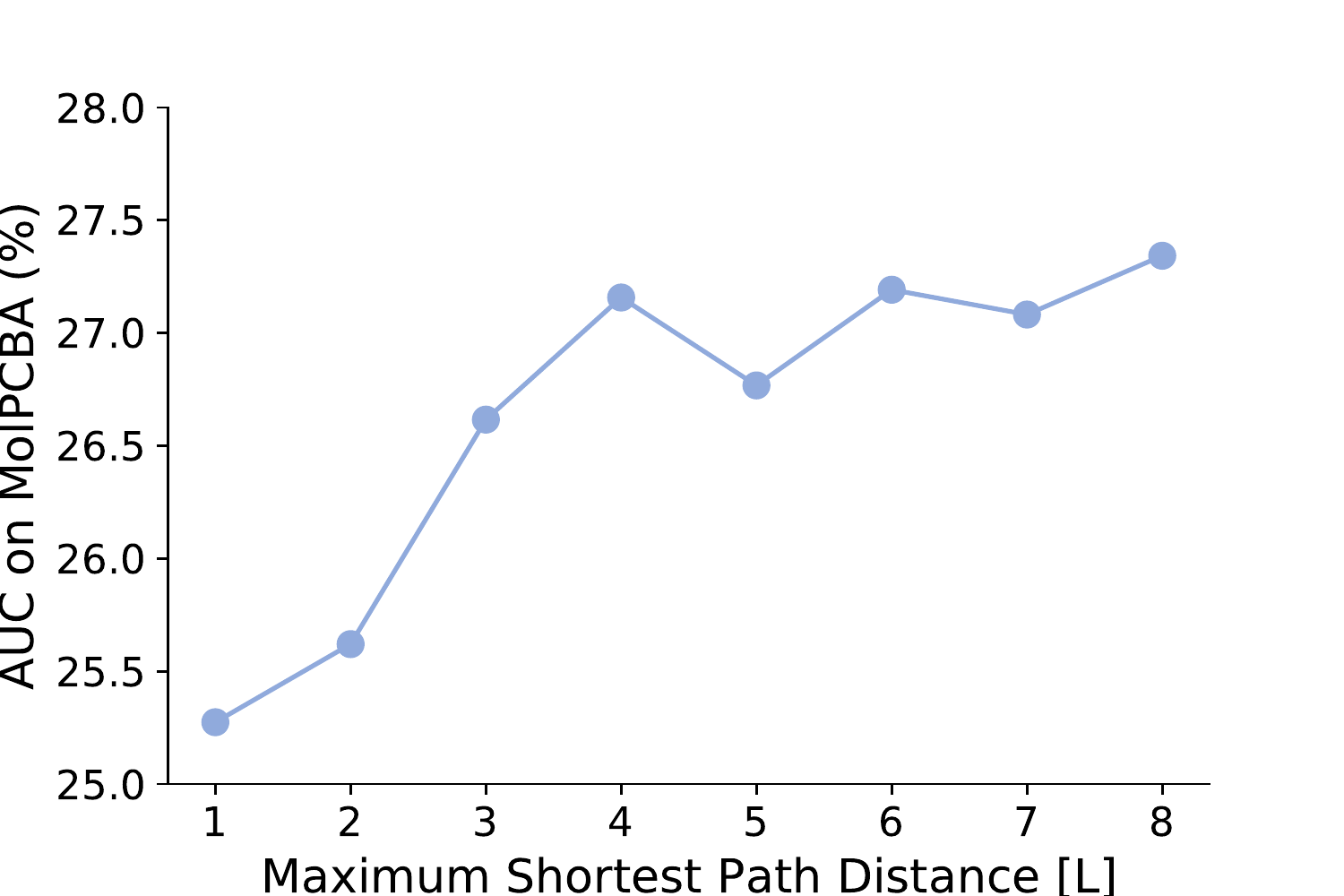}
\caption{Effects of the maximum shortest path distance $L$. The higher the better.}
\label{fig:abl_shortest}
\end{minipage}
\hfill
\end{figure}

We validate the effects of the maximum shortest path distance $L$.
% For the two nodes that are more than $L$-hop apart, we utilize a special structure embedding vector $\mathcal{P}_{\text{far}}$ to represent their structural relationship. % 양이넘치면 지울수도있음
We adopt GRPE-Standard, and the models are trained from scratch.
Figure~\ref{fig:abl_shortest} shows the ablation study result.
Increasing $L$ means that a model can identify the position of nodes that are further away.
The AUC consistently improves by increasing $L$ from one to four, but $L$ more than four does not further improve performance. 
% with distinctive structure embedding $\mathcal{P}_l$.
% However, the accuracy of the model drops after increase $L$ to more than four.
% Those results may be incurred by the increased number of parameters of structure embedding, which leads model to overfit on the training dataset.

\subsection{Sharing Topology Encoding and Edge Encoding}

We conduct ablation studies about the effects of sharing topology encoding $\mathcal{P}$ and edge encoding $\mathcal{E}$ for all layers. 
% Note that, $n$-th head of self-attention module has independent structure and edge embeddings.
We adopt GRPE-Small.
We conduct five independent runs.
The results in Table~\ref{tab:abl_sharing} show that sharing two encodings does improve MAE but not significantly.
% Empirically, sharing the encodings does not significantly change the performance of a model especially on large datasets.
% Those results may be incurred by the decreased number of parameters, which leads model not to overfit on the training dataset.
% However, empirically we found out that sharing the embeddings does not significantly changes the performance of a model on large datasets such as MolPCBA or PCQM4M dataset.

\begin{table}
\caption{Effects of sharing topology encoding and edge encoding for all layers. We report MAE on ZINC dataset. The lower the better.}
\centering
\vspace{-0.2cm}
\begin{tabular}{ccc}
\toprule
\multicolumn{1}{c|}{Is shared?} & \multicolumn{1}{c|}{\#Params} & MAE \\ \hline
\multicolumn{1}{c|}{yes}  & \multicolumn{1}{c|}{489k} & 
\multicolumn{1}{c}{\textbf{0.094}$\pm$ 0.002}\\
\multicolumn{1}{c|}{no}  & \multicolumn{1}{c|}{579k} & 
\multicolumn{1}{c}{0.101 $\pm$ 0.003} \\
 
\bottomrule
\end{tabular}
\label{tab:abl_sharing}
\end{table}

% \subsubsection{Effects of virtual node} We conduct ablation studies regarding the effects of virtual node on GRPE. 
% Virtual node is an alternative of global pooling layer where its feature represents an entire graph, which is useful for tasks such as graph classification or regression.
% Other than that, virtual node might be a useful bridge to transfer knowledge of entire graph to each node, which might be useful for tasks such as node classification.
% Therefore we conduct experiments on both ZINC for graph regression, and CLUSTER for node classification.
% We conduct five independent runs. We adopt GRPE-Small. 
% Table~\ref{tab:abl_vn_zinc} shows that without virtual node GRPE unable t

% \input{table/ablation_vn.tex}

\subsection{Attention Map of GRPE} We visualize attention map of each layer on molecules with various size. We take average over the attention map of 32 heads from GRPE-Standard trained on graph regression task on PCQM4Mv2 dataset.
Note that, the first node is virtual node (with index zero), and its attention map and adjacency is represented at the first row and column.
Figure~\ref{fig:viz_att} shows that our method attend to neighboring nodes at lower layers. 
As layer goes deeper, attention map attends to virtual node with higher attention on the first column.
This shows that each layer focuses on the entire graph rather than each node.

\section{Conclusion}
We studied the problem of positional encoding for representing structural information of the given graph better, specifically, by adding node-topology and node-edge relations to the model.
We validated the effectiveness of our approach, both quantitatively and qualitatively, in various tasks of diverse dataset sizes and characteristics, \eg, HIV replication prediction from molecule graphs (MolHIV) or semi-supervised clustering on synthetic graphs (CLUSTER).
% Notably, we achieved state-of-the-art results on molecule graphs property prediction tasks along with node classification, \eg, MolPCBA, MolHIV and  CLUSTER.

% From molecule graph property prediction tasks to node classification task, \eg, HIV replication prediction or semi-supervised clustering.

% Use \bibliography{yourbibfile} instead or the References section will not appear in your paper
\bibliography{aaai23}

\begin{thebibliography}{23}
\providecommand{\natexlab}[1]{#1}

\bibitem[{Beani et~al.(2021)Beani, Passaro, L{\'e}tourneau, Hamilton, Corso,
  and Li{\`o}}]{dgn_beani2021directional}
Beani, D.; Passaro, S.; L{\'e}tourneau, V.; Hamilton, W.; Corso, G.; and
  Li{\`o}, P. 2021.
\newblock Directional graph networks.
\newblock In \emph{International Conference on Machine Learning}, 748--758.
  PMLR.

\bibitem[{Belkin and Niyogi(2003)}]{belkin2003laplacian}
Belkin, M.; and Niyogi, P. 2003.
\newblock Laplacian eigenmaps for dimensionality reduction and data
  representation.
\newblock \emph{Neural computation}, 15(6): 1373--1396.

\bibitem[{Bresson and Laurent(2017)}]{gated_bresson2017residual}
Bresson, X.; and Laurent, T. 2017.
\newblock Residual gated graph convnets.
\newblock \emph{arXiv preprint arXiv:1711.07553}.

\bibitem[{Brossard, Frigo, and Dehaene(2020)}]{gine_brossard2020graph}
Brossard, R.; Frigo, O.; and Dehaene, D. 2020.
\newblock Graph convolutions that can finally model local structure.
\newblock \emph{arXiv preprint arXiv:2011.15069}.

\bibitem[{Cai et~al.(2021)Cai, Luo, Xu, He, Liu, and Wang}]{cai2021graphnorm}
Cai, T.; Luo, S.; Xu, K.; He, D.; Liu, T.-y.; and Wang, L. 2021.
\newblock Graphnorm: A principled approach to accelerating graph neural network
  training.
\newblock In \emph{International Conference on Machine Learning}, 1204--1215.
  PMLR.

\bibitem[{Corso et~al.(2020)Corso, Cavalleri, Beaini, Li{\`o}, and
  Veli{\v{c}}kovi{\'c}}]{pna_corso2020principal}
Corso, G.; Cavalleri, L.; Beaini, D.; Li{\`o}, P.; and Veli{\v{c}}kovi{\'c}, P.
  2020.
\newblock Principal neighbourhood aggregation for graph nets.
\newblock \emph{arXiv preprint arXiv:2004.05718}.

\bibitem[{Devlin et~al.(2018)Devlin, Chang, Lee, and
  Toutanova}]{devlin2018bert}
Devlin, J.; Chang, M.-W.; Lee, K.; and Toutanova, K. 2018.
\newblock Bert: Pre-training of deep bidirectional transformers for language
  understanding.
\newblock \emph{arXiv preprint arXiv:1810.04805}.

\bibitem[{Dwivedi and Bresson(2020)}]{gt_dwivedi2020generalization}
Dwivedi, V.~P.; and Bresson, X. 2020.
\newblock A generalization of transformer networks to graphs.
\newblock \emph{arXiv preprint arXiv:2012.09699}.

\bibitem[{Dwivedi et~al.(2020)Dwivedi, Joshi, Laurent, Bengio, and
  Bresson}]{dwivedi2020benchmarking}
Dwivedi, V.~P.; Joshi, C.~K.; Laurent, T.; Bengio, Y.; and Bresson, X. 2020.
\newblock Benchmarking graph neural networks.
\newblock \emph{arXiv preprint arXiv:2003.00982}.

\bibitem[{Gilmer et~al.(2017)Gilmer, Schoenholz, Riley, Vinyals, and
  Dahl}]{mpnn_gilmer2017neural}
Gilmer, J.; Schoenholz, S.~S.; Riley, P.~F.; Vinyals, O.; and Dahl, G.~E. 2017.
\newblock Neural message passing for quantum chemistry.
\newblock In \emph{International conference on machine learning}, 1263--1272.
  PMLR.

\bibitem[{He et~al.(2016)He, Zhang, Ren, and Sun}]{he2016deep}
He, K.; Zhang, X.; Ren, S.; and Sun, J. 2016.
\newblock Deep residual learning for image recognition.
\newblock In \emph{Proceedings of the IEEE conference on computer vision and
  pattern recognition}, 770--778.

\bibitem[{Hu et~al.(2020)Hu, Fey, Zitnik, Dong, Ren, Liu, Catasta, and
  Leskovec}]{hu2020open}
Hu, W.; Fey, M.; Zitnik, M.; Dong, Y.; Ren, H.; Liu, B.; Catasta, M.; and
  Leskovec, J. 2020.
\newblock Open graph benchmark: Datasets for machine learning on graphs.
\newblock \emph{arXiv preprint arXiv:2005.00687}.

\bibitem[{Hussain et~al.(2021)Hussain, Shamim, Zaki, J, and
  Subramanian}]{hussain2021edge}
Hussain; Shamim, M.; Zaki; J, M.; and Subramanian, D. 2021.
\newblock Edge-augmented graph transformers: Global self-attention is enough
  for graphs.
\newblock \emph{arXiv preprint arXiv:2108.03348}.

\bibitem[{Kipf and Welling(2016)}]{gcn_kipf2016semi}
Kipf, T.~N.; and Welling, M. 2016.
\newblock Semi-supervised classification with graph convolutional networks.
\newblock \emph{arXiv preprint arXiv:1609.02907}.

\bibitem[{Kreuzer et~al.(2021)Kreuzer, Beaini, Hamilton, L{\'e}tourneau, and
  Tossou}]{san_kreuzer2021rethinking}
Kreuzer, D.; Beaini, D.; Hamilton, W.~L.; L{\'e}tourneau, V.; and Tossou, P.
  2021.
\newblock Rethinking Graph Transformers with Spectral Attention.
\newblock \emph{arXiv preprint arXiv:2106.03893}.

\bibitem[{Le et~al.(2021)Le, Bertolini, No{\'e}, and
  Clevert}]{phc_le2021parameterized}
Le, T.; Bertolini, M.; No{\'e}, F.; and Clevert, D.-A. 2021.
\newblock Parameterized hypercomplex graph neural networks for graph
  classification.
\newblock \emph{arXiv preprint arXiv:2103.16584}.

\bibitem[{Li et~al.(2020)Li, Xiong, Thabet, and
  Ghanem}]{deepergn_li2020deepergcn}
Li, G.; Xiong, C.; Thabet, A.; and Ghanem, B. 2020.
\newblock Deepergcn: All you need to train deeper gcns.
\newblock \emph{arXiv preprint arXiv:2006.07739}.

\bibitem[{Rong et~al.(2020)Rong, Bian, Xu, Xie, Wei, Huang, and
  Huang}]{rong2020self}
Rong, Y.; Bian, Y.; Xu, T.; Xie, W.; Wei, Y.; Huang, W.; and Huang, J. 2020.
\newblock Self-supervised graph transformer on large-scale molecular data.
\newblock \emph{arXiv preprint arXiv:2007.02835}.

\bibitem[{Shaw, Uszkoreit, and Vaswani(2018)}]{shaw2018self}
Shaw, P.; Uszkoreit, J.; and Vaswani, A. 2018.
\newblock Self-attention with relative position representations.
\newblock \emph{arXiv preprint arXiv:1803.02155}.

\bibitem[{Vaswani et~al.(2017)Vaswani, Shazeer, Parmar, Uszkoreit, Jones,
  Gomez, Kaiser, and Polosukhin}]{vaswani2017attention}
Vaswani, A.; Shazeer, N.; Parmar, N.; Uszkoreit, J.; Jones, L.; Gomez, A.~N.;
  Kaiser, {\L}.; and Polosukhin, I. 2017.
\newblock Attention is all you need.
\newblock In \emph{Advances in neural information processing systems},
  5998--6008.

\bibitem[{Veli{\v{c}}kovi{\'c} et~al.(2017)Veli{\v{c}}kovi{\'c}, Cucurull,
  Casanova, Romero, Lio, and Bengio}]{velivckovic2017graph}
Veli{\v{c}}kovi{\'c}, P.; Cucurull, G.; Casanova, A.; Romero, A.; Lio, P.; and
  Bengio, Y. 2017.
\newblock Graph attention networks.
\newblock \emph{arXiv preprint arXiv:1710.10903}.

\bibitem[{Xu et~al.(2018)Xu, Hu, Leskovec, and Jegelka}]{gin_xu2018powerful}
Xu, K.; Hu, W.; Leskovec, J.; and Jegelka, S. 2018.
\newblock How powerful are graph neural networks?
\newblock \emph{arXiv preprint arXiv:1810.00826}.

\bibitem[{Ying et~al.(2021)Ying, Cai, Luo, Zheng, Ke, He, Shen, and
  Liu}]{ying2021transformers}
Ying, C.; Cai, T.; Luo, S.; Zheng, S.; Ke, G.; He, D.; Shen, Y.; and Liu, T.-Y.
  2021.
\newblock Do Transformers Really Perform Bad for Graph Representation?
\newblock \emph{arXiv preprint arXiv:2106.05234}.

\end{thebibliography}

\end{document}